\documentclass[lettersize,journal]{IEEEtran}
\usepackage{amsmath,amsfonts}
\usepackage{algorithmic}
\usepackage{algorithm}
\usepackage{array}
\usepackage[caption=false,font=normalsize,labelfont=sf,textfont=sf]{subfig}
\usepackage{textcomp}
\usepackage{stfloats}
\usepackage{url}
\usepackage{verbatim}
\usepackage{graphicx}
\usepackage{cite}
\usepackage{color}
\usepackage{subfig}
\usepackage[T1]{fontenc}

\begin{document}

\title{Fidelity-Induced Interpretable Policy Extraction for Reinforcement Learning}

\author{Liu Xiao, Chen Wubin, Tan Mao
\thanks{Liu Xiao and Tan Mao is with the School of Automation and Electronic Information, Xiangtan University, China.}
\thanks{Chen Wubin is with the Department of Computer Science and Technology, Nanjing University, China.}
}

\markboth{Journal of \LaTeX\ Class Files,~Vol.~14, No.~8, August~2021}%
{Shell \MakeLowercase{\textit{et al.}}: A Sample Article Using IEEEtran.cls for IEEE Journals}


\maketitle

\begin{abstract}
Deep Reinforcement Learning (DRL) has achieved remarkable success in sequential decision-making problems. However, existing DRL agents make decisions in an opaque fashion, hindering the user from establishing trust and scrutinizing weaknesses of the agents. 
While recent research has developed Interpretable Policy Extraction (IPE) methods for explaining how an agent takes actions, their explanations are often inconsistent with the agent's behavior and thus, frequently fail to explain.
To tackle this issue, we propose a novel method, Fidelity-Induced Policy Extraction (FIPE). Specifically, we start by analyzing the optimization mechanism of existing IPE methods, elaborating on the issue of ignoring consistency while increasing cumulative rewards. We then design a fidelity-induced mechanism by integrate a fidelity measurement into the reinforcement learning feedback.
We conduct experiments in the complex control environment of StarCraft II, an arena typically avoided by current IPE methods. The experiment results demonstrate that FIPE outperforms the baselines in terms of interaction performance and consistency, meanwhile easy to understand.
\end{abstract}


\section{Introduction}
\IEEEPARstart{D}{eep} Reinforcement Learning (DRL) has showcased impressive achievements in various sequential decision-making domains. For example, it has been employed to train agents that outperform professional players in complex games~\cite{molnar2020interpretable,vinyals2019grandmaster,lanctot2017unified,silver2017mastering} and control robots for intricate tasks~\cite{collins2005efficient,johannink2019residual,li2023deep}. Despite its capacity to learn complex mappings between inputs and outputs, DRL represent these knowledge in an opaque fashion. This lack of transparency hides the key factors influencing decision-making and exacerbates the difficulty of identifying and rectifying errors. \cite{bastani2016measuring,guo2021edge,romdhana2022deep,minh2022explainable,loh2022application}. Consequently, the practical application of DRL methods is limited, particularly in cost-sensitive domains like autonomous driving, healthcare, and finance.

\begin{figure*}
    \centering
    \subfloat[]{\includegraphics[width=.5\linewidth]{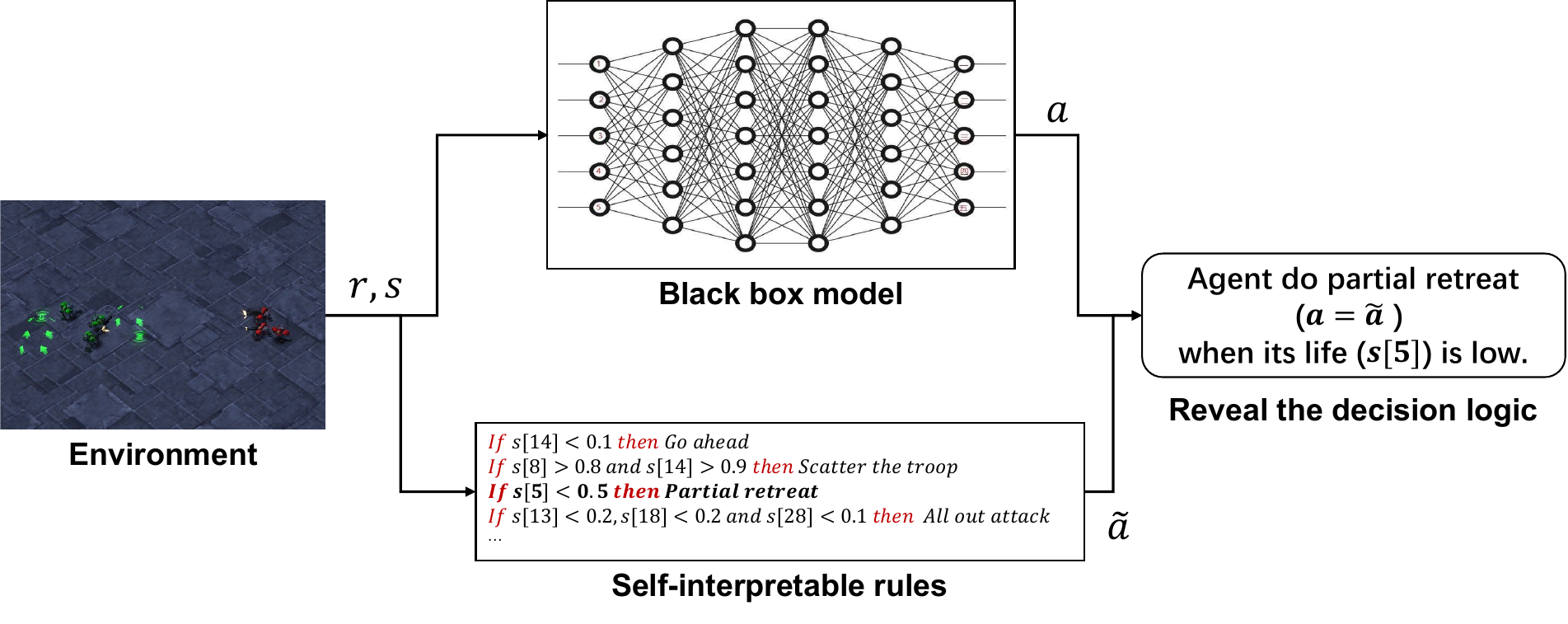}}
    \subfloat[]{\includegraphics[width=.5\linewidth]{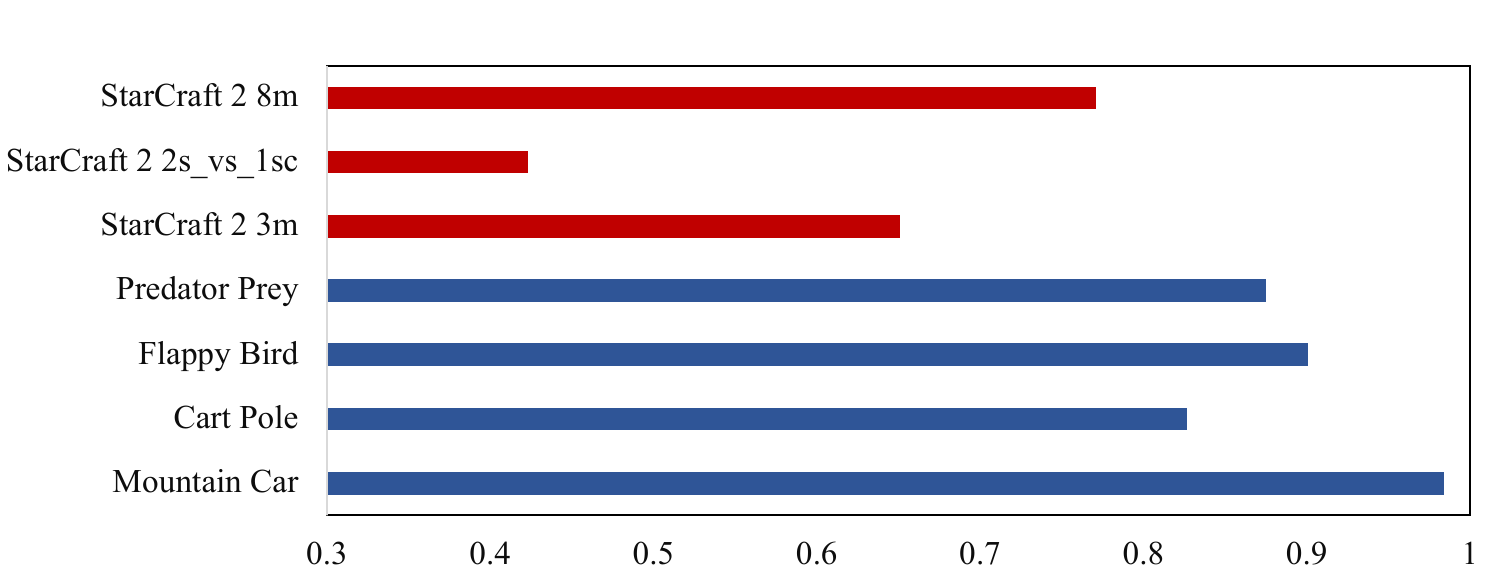}}
    \caption{(a) Explain an agent's behavior using self-interpretable rules. (b) Success rate of explanation using existing policy extraction methods (using VIPER~\cite{bastani2018verifiable} as an example). The vertical axis represents the tasks, and the horizontal axis represents the success rate of explanations.
}
    \label{fig.explanation_failure}
\end{figure*}

To address the problem of opaque knowledge representation, previous research ~\cite{ross2011reduction,lee2019complementary, dahlin2020designing, bastani2018verifiable, liu2019rethinking, li2021neural} propose to extract self-interpretable policy from the interactive tranjections of DRL policy. At a high level, these approaches leverage the transparent structure of self-interpretable models to unveil the hidden knowledge within DRL systems (see Figure~\ref{fig.explanation_failure} (a)). Specifically, the self-interpretable models are trained based on the interaction data, consisting of state-action pairs, collected during the interaction between agents and their environment. When the self-interpretable model is provided with the same state as the agent, it reveals the key factors by showcasing the rules that are activated within the self-interpretable model.
However, despite their ability to identify key features in specific observations, these methods suffer from a problem of \textbf{inconsistent explanation}. Specifically, the goal of interpretable policy extraction is to describe the DRL's decisions accurately with an interpretable structure. When the extracted policy diverges from the decisions made by the DRL policy, the output explanations fail to interpret the agent's decisions.
According to Figure~\ref{fig.explanation_failure} (b), the consistency of the current IPE policies is affected by the difficulty of the task. A typical example is that, when dealing with the $2s\_vs\_1sc$ task in StarCraft II, the success rate of existing method dropped to about 40\%. 
In conclusion, existing IPE methods aim to maximize rewards rather than the fidelity of the extracted policies throughout the entire lifecycle, which deviates from the goal of explaining the DRL policy, particularly in complex tasks like those in the StarCraft II platform.

This paper proposes a novel method called Fidelity-Induced Policy Extraction (FIPE) to address the problem of inconsistent explanation. At a high level, FIPE introduces a fidelity-induced mechanism that integrates the fidelity metric into the reinforcement learning feedback and thus, guiding the model toward improving its consistency with the teacher. Specifically, we first analyze the preference towards higher rewards exhibited by current methods from a theoretical perspective, highlighting their deviation from the original intent of interpretable policy extraction. Subsequently, we design a fidelity-induced mechanism to rectify this preference, and estimate the upper limits of this mechanism. Finally, we propose an approximate solution for the fidelity-induced mechanism, simplifying the calculation of the FIPE mechanism, especially in complex tasks.

We conduct experiments to evaluate the proposed method within three complex multi-agent reinforcement learning tasks in the StarCraft II platform, i.e., $3m$, $2s\_vs\_1sc$, $8m$. According to the experiment results, the proposed method outperforms the baselines in interactive performance and consistency. Furthermore, we demonstrate use cases for the FIPE compatible with other self-explainable structures. Overall, the proposed method exhibits greater applicability in complex tasks.

\section{Related Works}
Interpretable Policy Extraction (IPE) is based on the concept of imitation learning~\cite{hussein2017imitation,ghasemipour2020divergence,cha2020proxy}.
At a high level, IPE aims to mimic the behavior of the DRL agents by modeling the interaction trajectories using a self-explainable model structure.

Some works propose novel interpretable structures to replace deep neural networks\cite{liu2019toward, vasic2022moet, coppens2019distilling}. The primary advantage of these emerging structures lies in their inherent interpretability, which summarizes the decision logic of the DRL agent. However, it is challenging to balance accuracy and interpretability in a model, i.e., the accuracy-interpretability trade-off~\cite{you2022interpretability, tjoa2020survey, puiutta2020explainable, rosenfeld2019explainability, silva2020optimization}. Empirically, more flexible methods that are capable of estimating more complex shapes for the unknown function are way less interpretable. Consequently, these innovative model structures are often more complex than traditional interpretable structures, making them more difficult to understand.

Some works~\cite{lee2019complementary, dahlin2020designing, guo2023explainable, wang2023explainable} improve the traditional interpretable model structure. This approach avoid to use uninteretable components and thus, provides concise explanations for the decison of DRL agents. Nevertheless, traditional interpretable structures (e.g., decision trees) struggle to generalize to complex tasks due to their unstable performance.

Recent studies have shifted their focus towards optimizing the extraction process
~\cite{ross2011reduction, bastani2018verifiable, liu2019toward, li2021neural, milani2022maviper, liu2023effective}. These methods involve selectively retaining interactive data that is deemed more ``significant'' for generating self-explainable outputs. This process improve the performance of self-explainable rule-based models. While they perform well in simple environments (e.g., Gym), these methods encounter issues of inconsistent explanation when deployed in challenging tasks (e.g., StarCraft II) due to the disparity between interpretable and original policies (refer to Figure~\ref{fig.explanation_failure} (b)).

\section{Preliminary}
In order to provide interpretable representations of the ``black-box'' decision logic of deep reinforcement learning models. At a high level, existing policy extraction methods employ a supervised learning paradigm to fit the sampled data of the model, i.e., state, action, and $Q$-value, denoted as $(s, a, q)$. This process transforms the model into an interpretable sequence of rules. Technically, existing approaches implicitly extract high-quality samples from the sampled data, enabling the extracted rule-based policies to achieve greater rewards within the environment. To elaborate on this process, we discuss the feedback and optimization mechanisms of reinforcement learning from the standpoint of Interpretable Policy Extraction. Table \ref{c5:preliminaries_symbols} summarizes the relevant symbols and variables of this paper.

\begin{table*}[htb]
\centering
\caption{Symbols and Variables Used in this Chapter}
\label{c5:preliminaries_symbols}
\begin{tabular}{ll}
\hline
\textbf{Symbol/Variable}                           & \textbf{Meaning}                       \\ \hline
$S, A, P, R$                                       & State set, action set, state transition probabilities, and reward function, respectively. \\ \hline
$U$                                              & Discount return, or cumulative discont reward.        \\ \hline
$\pi, \pi^*$                                              & DRL policy and its optimized policy.        \\ \hline
$\tilde{\pi}, \tilde{\pi}^*$                                              & Extracted interpretable policy and its optimized policy.        \\ \hline
$V,Q,\gamma$                                              & state value, action value, discount factor.        \\ \hline
$C_s, J$                                              & Cost of extracted policy at a specific state, cost function. \\ \hline
$d^\pi,\tau$                                             & State distribution, interaction trajectory.                  \\ \hline
$\aleph$                                       & Fidelity-induced component.          \\ \hline
\end{tabular}
\end{table*}

\subsection{Implicit Knowledge in Reinforcement Learning Feedback}
To maximize the cumulative rewards obtained from the interaction between an agent and its environment, an action-reward function is used to guide the intelligent agent in learning the optimal policy. When sampling with a well-trained policy, the $Q$-value in the interaction samples reflects the expectation of discounted rewards for a specific action.

Technically, in a finite-horizon Markov Decision Process (MDP) defined by the tuple $(S, A, P, R)$, where $S$ represents the state set, $A$ represents the action set, $P:S\times A\times S\rightarrow[0,1]$ (i.e., $P(s,a,s')=p(s'|s,a)$) represents the environment transition probability function, and $R:S\rightarrow\mathbb{R}$ represents the reward function, we consider that rewards closer to the current time are more valuable than future rewards, which may not materialize. Therefore, we often use the discounted return to express a preference for future expected returns, denoted as $U=\mathbb{E}[\sum_{k=t+1}^{+\infty}\gamma^{k-t-1}r_k]$, where $\gamma\in [0,1]$ is the discount factor. Smaller values of $\gamma$ assign less importance to future rewards. Using the discounted reward, we can define the action value at time $t$, representing the quality of action $a_t$ in state $s_t$ and allowing us to evaluate actions. Specifically, the action value function (also known as the $Q$ function) is formulated as
\begin{equation}
    Q_\pi(s_t,a_t)=\mathbb E[U_t|S_t=s_t,A_t=a_t].
\end{equation}

Current policy extraction methods leverage supervised learning to train models using sampled data, which includes states, actions, and $Q$-values, denoted as $(s, a, q)$. Among these, the $Q$-value signifies the anticipated discounted reward for a given sample. As a result, policy extraction methods enhance the performance of interpretable rule-based policies in tasks by preserving diverse samples with high expected rewards.

\subsection{Optimization Approach for Interpretable Policy Extraction}
Interpretable Policy Extraction aims to distill rule based policy that accurately represent the behavioral of the original agent's policy~\cite{ross2011reduction,bastani2018verifiable}. Technically, given a teacher policy model $\pi^*$ and a student policy model $\tilde{\pi}$, for a specific state $s$, the extracted rule-based policy selects action $a$, incurring a cost to the teacher policy $\pi^*$ denoted as $C_s (\pi^*,\tilde{\pi})$. Generally, the $C_s (\pi^*,\tilde{\pi})$ is adaptable, and different definition of $C_s (\pi^*,\tilde{\pi})$ result in policies with different preferences. Typically, policy extraction methods optimize losses based on rewards, denoted as $l(s,\tilde\pi)$. We have:
\begin{align}
    C_s (\pi^*,\tilde{\pi})
     & =\mathbb E_{a\sim\tilde\pi(s)}[C(s,\tilde\pi)]  \\
     & =\mathbb E_{a\sim\tilde\pi(s)}[l(s,\tilde\pi)].
\end{align}
At this point, the overall cost-to-go function of $\tilde{\pi}$ with respect to $\pi$ at time step $T$ is expressed as:
\begin{align}
    J(\tilde\pi)
     & =\sum_{t=1}^T\mathbb{E}_{s\sim d^{\tilde\pi}} [C_s(\pi^*,\tilde\pi)] \\
     & =T\mathbb E_{s\sim d^{\tilde\pi}}[C_s(\pi^*,\tilde\pi)]              \\
     & =T\mathbb{E}_{s\sim d^\pi}[l(s,\tilde\pi)].
\end{align}
To address the issue of accumulating sequential decision errors~\cite{ross2011reduction}, a common approach involves using a mixed strategy between the student and teacher to generate interaction trajectories. The mixed strategy is represented as:
\begin{equation}
    \pi_i = \beta \pi^* + (1-\beta) \tilde\pi,
\end{equation}
where $\pi_i$ represents the mixed strategy for the $i$-th round of interaction, and $\beta\in[0,1]$ is a decreasing parameter. As a result, the interaction trajectory obtained in each round is denoted as $\tau = \{(s_t, \pi_i(s_t))\}_{t=1:T}$. The goal of policy extraction is to find a student policy $\tilde\pi^*$ that minimizes the average cost under the induced state distribution, formally represented as:
\begin{align}
    \tilde \pi^*
     & =\arg\min_{\tilde\pi}\mathbb E_{s\sim d^{\tilde\pi}} (J(\tilde\pi)(\pi^*,\tilde \pi)) \\
     & =\arg\min_{\tilde\pi}T\mathbb E_{s\sim d^{\tilde\pi}}[l(s,\tilde\pi)]                 \\
     & =\arg\min_{\tilde\pi}\mathbb E_{s\sim d^{\tilde\pi}}[l(s,\tilde\pi)].
\end{align}
The student policy $\tilde\pi$ is typically represented with structured and semantically clear rules, e.g., decision trees or nearest neighbor models, making it easy to understand.

\section{Analyzing Optimization Preferences of Current Policy Extraction Methods}
Current policy extraction methods learn interpretable rule-based policy from deep reinforcement learning policies. However, they often fall short in addressing the challenge of inconsistent explanation. Specifically, these interpretable rule-based policies can sometimes produce outcomes that do not align with the original policies. When such inconsistencies happen, the explanations provided by these methods lose their effectiveness. To deal with this issue, we first analyze the optimization mechanisms employed by current policy extraction methods.

Consider a well-trained deep reinforcement learning policy, denoted as $\pi^*:S\rightarrow A$, where $\pi^*$ maps states $S$ to actions $A$. At any time step $t$ within the set $\{0,...,T-1\}$, the agent's policy $\pi^*$ selects an action $a_t\in A$ that maximizes the expected average reward when given an input state $s_t\in S$. For any policy $\pi$, we denote the state distribution at time step $t$ as $d_\pi^t$. When the initial state is $s_0\in S$, the state distribution $d^\pi$ over time steps $t=0:T$ is formalized as:
\begin{equation}
    \begin{cases}
        d_0^\pi=\mathbb{I}(s=s_0),\quad (t=0) \\
        d_t^\pi=\sum_{s\in S}P(s,\pi(s),s')d_{t-1}^\pi(s),\quad (t>0),
    \end{cases}
\end{equation}
where $\mathbb{I}$ represents the indicator function. When the agent interacts with the environment using policy $\tilde\pi$, it generates an interaction trajectory $\tau=\{(s_t,\tilde\pi(s_t))\}_{t=1:T, s_t\sim d^{\tilde\pi}}$.

As for the Q-DAGGER algorithm~\cite{bastani2018verifiable}, the state-value function and action-value function are formalized as
\begin{align}
    V_t^{\pi}(s)   & = R(s) + \sum_{s' \in S} P(s, \pi(s), s')V_{t+1}^\pi(s'), \\
    Q_t^{\pi}(s, a) & = R(s) + \sum_{s' \in S} P(s, a, s')V_{t+1}^\pi(s').
\end{align}
Therefore, the policy extraction loss for a specific state $s$ is defined as:
\begin{align}
    l(s, \tilde\pi) = V_t^{\pi^*}(s) - Q_t^{\pi^*}(s).
\end{align}
Let $g(s, \tilde\pi) = \mathbb{I}[\tilde\pi(s) \neq \pi^*(s)]$ be the 0-1 loss, with its upper bound denoted as $\tilde g(s, \tilde\pi)$. The upper bound of $l(s, \tilde\pi)$ is given by
\begin{align}
    \tilde l(s, \tilde\pi)
    & = \tilde g(s, \tilde\pi)\tilde l_t(s) \\
    & = [V_t^{\pi^*}(s) - \min_{a \in A}Q_t^{\pi^*}(s,a)]g(s, \tilde\pi),
\end{align}
where $\min_{a \in A}$ finds the minimum over the action space $A$. $g(s, \tilde\pi) = \mathbb{I}[\tilde\pi(s) \neq \pi^*(s)]$ considers both the extracted policy $\tilde\pi$ and the optimal teacher policy $\pi^*(s)$. This method does not explicitly measure similarity but focuses on selecting advantageous samples for state $s$. Specifically, given a state $s$, when $g(s, \tilde\pi) = 0$, it indicates that the two policies are alike, and according to the definition of $\tilde l(s, \tilde\pi)$, their loss is zero:
\begin{equation}
    l_t(s)g(s, \tilde\pi) = l(s, \tilde\pi) = 0.
\end{equation}
When $g(s, \tilde\pi) = 1$, we have:
\begin{equation}
    l_t(s)g(s,\tilde\pi)=l(s,\tilde\pi)=V_t^{\pi^*}(s)-\min_{a\in A}Q_t^{\pi^*}(s,a).
\end{equation}
At this point, the loss function represents the difference between the action value $Q$ for a specific action and the average action value $V$ for that state. At a high level, this method prefers actions that hold an advantage in the current state. Minimizing this loss will increase the probability of selecting advantageous samples and thus, induce the extracted interpretable policy that prefers cumulative reward. However, there are two issues that must be considered: First, this approach does not explicitly model fidelity to the original policy; Second, the interpretable model structure may not precisely represent the optimal policy. As a result, existing IPE methods suffer from a bottleneck of inconsistent explanation. When such inconsistent explanation occur, the policy explanations lose their effectiveness.

\section{Theoretical Insights of The Fidelity-Induced Mechanism}\label{section_theory}
We propose Fidelity-Induced Mechanism to tackle the problem of inconsistent explanation. In this section, we provide a theoretical analysis of the feasibility of the fidelity-induced mechanism.

\subsection{Hypotheses}
We present two hypotheses that underlie the fidelity-induced mechanism together with other IPE methods. On one hand, since the black-box system cannot be fully disclosed during, it is important to validate the accuracy of the black-box system representation. On the other hand, the black-box system must be representable.

\subsubsection{Policy Equivalence Hypothesis}
The policy equivalence hypothesis supports the correctness of the policy extraction results. Generally, IPE involves two systems, denoted as $F_1$ and $F_2$, engaging in an interactive task $T$. Both systems independently interact with task $T$ under the same initial state, sample interaction trajectories $\tau_{F_1}$ and $\tau_{F_2}$. Within a finite number of interactions, if the interaction trajectories $\tau_{F_1}$ and $\tau_{F_2}$ are consistent, then systems $F_1$ and $F_2$ are equivalent.

However, this hypothesis overlooks certain complexities:
(a) From the perspective of logical consistency, equivalence between two systems includes both process and outcome. Therefore, achieving consistent output results is not a sufficient and necessary condition for system equivalence. Given the unknown internal logic of the black-box system, assessing the system's processes is impossible. This hypothesis primarily focuses on result consistency, indirectly implying system consistency.
(b) From the perspective of verification, given the inability to unveil the black-box system, it is essential to determine the consistency of two systems through extensive verification. However, in practical applications, conducting extensive verification is often impractical, necessitating the use of finite validation as a substitute. This hypothesis takes into consideration the practicality of verification by translating finite interaction consistency into an approximation of consistency when dealing with infinite interactions in the context of reinforcement learning.

\subsubsection{Stability Hypothesis}
The stability hypothesis plays a pivotal role in bolstering the viability of policies obtained through IPE. Specifically, in the context of two systems denoted as $F_1$ and $F_2$, where $F_1$ represents a transparent system and $F_2$ signifies an opaque policy. On the premise that $F_2$ and $T$ do not change, the objective of $F_1$ is to mimic the behavior of $F_2$, while subsequently verifying this emulation through interaction with task $T$. However, it is important to note that the stability hypothesis may to some extent overlook potential fluctuations inherent in both the black-box system and the interactive task.

Specifically, this hypothesis overlooks certain complexities:
(a) When it comes to the black-box policy, exemplified in the realm of deep reinforcement learning, the policy undergoes continual updates during interactions, and each interaction has the potential to induce modifications in the policy itself. As a result, the stability hypothesis fails to consider the potential alterations that can occur in deep reinforcement learning policies. This oversight can be mitigated when the policy is fully trained or when its parameters are fixed artificially to prevent further changes.
(b) Regarding the interactive task, the stability hypothesis presupposes that the task used for validation remains constant over time and across interactions. In the traditional RL framework, this hypothesis disregards the possibility of task to change during interactions. In other words, the hypothesis implies that, for any given input, the task consistently produces an output probability distribution. It's crucial to note that internal randomness within the output does not undermine task stability, as the cumulative effect of repeated interactions ensures that the overall output probability distribution remains unchanged.

\subsection{Design of the Fidelity-Induced Mechanism}
This section expounds upon the structural framework of the Fidelity-Induced Policy Extraction (FIPE) method. At a high level, FIPE replaces the existing reward-centric optimization objective in policy extraction methods with a fidelity-induced loss function designed to prioritize fidelity.

To delve deeper into this concept, within the realm of a finite-horizon Markov Decision Process (MDP), the fully optimized black-box policy is denoted as $\pi^*$. The policy extraction methodology constructs the loss function in the following manner:
\begin{align}
    C_s(\pi^{*},\tilde\pi)
     & =\mathbb E_{a\sim\tilde\pi(s)}[C[s,\tilde\pi]]\\
     & =\mathbb E_{a\sim\tilde\pi(s)}[l(s,\tilde\pi)]\\
     & =\mathbb E_{a\sim\tilde\pi(s)}[V_t^{\pi^{*}}(s)-Q_t^{\pi^{*}}(s)]\\
     & =\mathbb E_{a\sim\tilde\pi(s)}[R(s)+\sum_{s'\in S}P(s,\pi^{*}(s),s')V_{t+1}^{\pi^{*}}(s')\\
     & -(R(s)+\sum_{s'\in S}P(s,a,s')V_{t+1}^{\pi^{*}}(s'))]\\
     & =\mathbb E_{a\sim\tilde\pi(s)}[V_{t+1}^{\pi^{*}}\sum_{s'\in S}(P(s,\pi^{*}(s)-P(s,a,s')))].
\end{align}
Where $V$ and $Q$ represent the state-value and action-value functions, respectively. which capture the agent's expected rewards, considering both state transitions and action choices. The policy extraction procedure begins with data collection from the environment, followed by fitting the posterior distribution of this data using an interpretable policy structure. This sequence of events follows a clear chronological order, implying that optimization occurs after a single round of interaction or upon reaching a time horizon of $T$ steps.

By introducing a fidelity loss component into the reward function, we make it possible to explicitly incorporate fidelity-induced optimization objectives, which in turn lead to fidelity-induced policy evaluations. We define the fidelity-induced reward as follows:
\begin{align}
    R_t^\aleph(s, \pi^*, \tilde\pi) = \left[U_t(s) - \eta\left(\pi(a|s;\theta) - \tilde\pi(a|s)\right)^2\right]_{s\sim d_{t=1:T}^\pi(s_0)},
\end{align}
Where $U_t(s) = \mathbb{E}[\sum_{k=t+1}^{+\infty}\gamma^{k-t-1}r_k]$ represents the cumulative discounted return starting from time step $t$ for a single interaction episode lasting $T$ time steps. The parameter $\eta$ denotes the fidelity coefficient, with a higher value emphasizing fidelity and a lower value focusing on the overall return. Consequently, the loss function for fidelity-induced policy extraction is formulated as
\begin{align}
    &E_{a\sim\tilde\pi(s)}[l^\aleph(s,\pi^*,\tilde\pi)] \\
    &= \mathbb E_{a\sim\tilde\pi(s)}\left[V_t^{\aleph}(s)-Q_t^{\aleph}(s)\right] \\
    &= \mathbb E_{a\sim\tilde\pi(s)}
    [R_t^\aleph+\sum_{s'\in S}P(s,\pi^*(s),s')V_{t+1}^\aleph(s')\\
    &\quad-(R_t^\aleph+\sum_{s'\in S}P(s,a,s')V_{t+1}^\aleph].
\end{align}
In the expression $R^\aleph(s, \tilde\pi, \pi^*)$, the fidelity-induced optimization objective is explicitly represented by $\eta[\pi(a|s;\theta) - \tilde\pi(a|s)]^2$, which in turn influences $C_s^\aleph$, ultimately leading to the derivation of interpretable policies with a heightened emphasis on fidelity. Remarkably, this adjustment does not have an impact on policy return as the sample size approaches infinity. In particular, as the sample size increases, the interpretable policy theoretically converges towards deep reinforcement learning policy, denoted as
\begin{align}
    \mathcal{L}_{tr} &= \mathcal{L}_{val} \\
    \sum_{(s, \pi(s))\in E} \frac{1}{\sum_{m=0}^{M} T_m} \Psi &= \sum_{s\in S} P(s| \mathcal{X}) \Psi,
\end{align}
Where $\Psi=\mathbb{I}(\pi(s)=\hat{\pi}(s))$. As a result, we have $\eta[\pi(a|s;\theta)-\tilde\pi(a|s)]^2 \rightarrow 0$. However, due to the complexity of solving this problem, this study introduces an approximate representation to address this issue.

\subsection{Theoretical Analysis of Fidelity-Induced Policy Extraction Optimization}

In this section, we discuss the optimization of fidelity-induced policy extraction.

\textbf{Theorem}:
Consider the objective function $J(\pi) = -V_0^{(\pi)}(s_0)$. If the new policy satisfies the condition:
\begin{align}
\mathbb{E}_{s\sim d^{\pi^{*}}}[\sum_{s'\in S}(P(s,\pi^{*}(s),s')-P(s,a,s'))V_{t+1}^{\pi^{*}}]\geq 0,
\end{align}
then the new policy will not perform worse than the original one, and the quantity $U^{\pi^{*}} - U^\pi \geq 0$ can be approximately optimized.

\textbf{Proof}:
By employing the policy improvement theorem, let $\pi'$ be an improvement over $\pi$. If, for all states $s\in S$, the inequality $Q^{\pi'}(s)\geq V^{\pi}(s)$ holds, then policy $\pi'$ will not perform worse than the original policy. 
Building upon the work presented in \cite{bastani2018verifiable}, we have the following expressions:
\begin{small}\begin{align}
    Tl & = J(\pi) - J(\pi') \\
            & = V_0^{\pi'} - V_0^{\pi} \\
            & = R(s) + \sum_{s'\in S}P(s,\pi'(s),s')V_{t+1}^{\pi'}(s') - R(s) \\
            & - \sum_{s'\in S}P(s,\pi(s),s')V_{t+1}^{\pi}(s') \\
            & = \sum_{s'\in S}P(s,\pi'(s),s')V_{t+1}^{\pi'}(s') - \sum_{s'\in S}P(s,\pi(s),s')V_{t+1}^{\pi}(s') \\
    l & = \frac{1}{T}\sum_{s'\in S}P(s,\pi'(s),s')V_{t+1}^{\pi'}(s') - P(s,\pi(s),s')V_{t+1}^{\pi}(s').
\end{align}\end{small}
In accordance with the previous definition, policy extraction yields the advantage of the sample as $l=V^{\pi}(s)-\min_{a\in A}Q^{\pi}(s,a)$~\cite{bastani2018verifiable}. Since Q and V are samples from the original reinforcement learning policy, we have $Q^{\pi'}(s)=Q^{\pi}(s)$. Then we have
\begin{align}
    V^{\pi}(s)-Q^{\pi'}(s)=V^{\pi}(s)-Q^{\pi}(s)\leq V^{\pi}(s)-\min_{a\in A}Q^{\pi}(s,a)
\end{align}

Therefore, when the condition $l\leq 0$ is met, the policy $\pi'$ will not perform worse than the original policy, i.e., $V^{\pi}(s)-Q^{\pi'}(s)\geq 0$. We calculate that the expectation of $l$ at a finite time step $T=N$, when
\begin{align}
    l\approx\sum_{t}^{t+N}\mathbb E_{s\sim d_t^\pi}[V^{\pi}(s)-\min_{a\in A}Q^{\pi}(s,a)]\leq 0
\end{align}
is met, the new policy will not perform worse than the original one. This expression can be further reformulated as follows:
\begin{small}\begin{align}
    & \sum_{t=0}^{T-1}\mathbb E_{s\sim d_t^\pi}[V^{\pi}(s)-\min_{a\in A}Q^{\pi}(s,a)]\leq 0 \\
    & \sum_{t=0}^{T-1}\mathbb E_{s\sim d_t^\pi}[l_t(s,\pi)]\leq 0 \\
    & \sum_{t=0}^{T-1}\mathbb E_{s\sim d_t^{(\pi)}}[\sum_{s'\in S}(P(s,\pi(s),s')V_{t+1}^{\pi}(s')\\
    &\quad -P(s,a,s')V_{t+1}^{\pi}(s'))]\leq 0 \\
    & \sum_{t=0}^{T-1}\mathbb E_{s\sim d_t^{(\pi)}}[\sum_{s'\in S}(P(s,\pi(s),s')-P(s,a,s'))V_{t+1}^{\pi}(s')]\leq 0 \\
    & T\mathbb E_{s\sim d^{(\pi)}}[\sum_{s'\in S}(P(s,\pi(s),s')-P(s,a,s'))V_{t+1}^{\pi}(s')]\leq 0.
\end{align}\end{small}
Therefore, having $V_{t+1}^{\pi}(s')$ and $P(s,\pi(s),s')-P(s,a,s')$ exhibit the same positive or negative sign is a sufficient but not necessary condition to satisfy the inequality. Intuitively, to optimize the policy, we should increase the probability of encountering samples with advantages and decrease the probability of encountering samples with low rewards. In the context of policy iteration, data is usually collected first, followed by iterative updates. This implies that optimization occurs after one round of interaction or after sampling for $T$ time steps. Therefore, the return of $T$ steps, i.e., $U^{(\pi^*)}-U^{(\pi)}$, can be regarded as an unbiased estimator of average return, i.e., $\sum_{s'\in S}(P(s,\pi(s),s')-P(s,a,s'))V_{t+1}^{\pi}(s')$, under the distribution $d^{(\pi)}_{i=t:T}(s_0)$. 

In order to make the model can be updated with the data of any $T=N$ time steps, it is feasible to approximate $U$ by obtaining the total return for $T=N$ steps, denoted as $U\approx \sum_{i=t}^N{R_i}$, where $R_i$ represent the actual reward at a specific time step. Consequently, to enhance the performance of the new policy relative to the original one or to ensure $Q^{\pi'}(s)\geq V^{\pi}(s)$, we estimate rewards from one round of sampling to increase the probability of encountering advantageous sampling sequences in the interaction experience pool and decrease the probability of encountering low-reward sampling sequences.

\section{Approximate Implementation of Fidelity-Induced Policy Extraction}
In this section, we outline an approximate implementation of the Fidelity-Induced Policy Extraction (FIPE) framework. At a high level, our goal is to estimate the advantage of a policy $\pi^*$ over another policy $\pi$ by considering the cumulative return over a fixed number of steps ($N$). We achieve this by retaining interaction experiences that enhance fidelity and iteratively update the rule-based model. Subsequently, to update the rule-based model, we devise a competitive mechanism to preserve high-quality rules during iterations.

\subsection{Fidelity-Induced Rule Optimization}
To optimize the distilled rules, we introduce a fidelity-induced feedback mechanism to preserve advantageous interaction experiences.

First, to mitigate the problem of error accumulation in sequential decision-making, as discussed in Ross et al.~\cite{ross2011reduction}, we employ a combination of the student-teacher strategy to sample the environment. This mixed strategy is represented as:
\begin{equation}
    \pi_i = \beta \pi^* + (1-\beta) \tilde\pi
\end{equation}

Next, to estimate the benefits of the current strategy, we incorporate a fidelity-induced term into the reward function. This ensures that policy evaluation takes into account the consistency of the policy. For each episode, we estimate the value of the interaction data sampled under the current strategy. To enhance the consistency of the policy extraction model, we calculate this value using fidelity-inducing mechanisms, denoted as $U = \sum_{t=1}^{T}{C_{s_t}^\aleph}$.

Finally, based on the original policy interaction dataset and the new interaction dataset, we construct two policies, $\tilde\pi$ and $\tilde\pi'$. We estimate the advantage $U^{(\pi^*)}-U^{(\pi)}$ based on the cumulative return values over $N$ steps. Specifically, when $U^{\pi'} - U^{\pi} > 0$, indicating that the new policy outperforms the original one, we choose to retain the new policy. Conversely, when $U^{\pi'} - U^{\pi} < 0$, we keep the existing policy.

\subsection{Policy Competition}\label{PolicyCompetition}
To prevent the issue of overfitting interpretable rule-based policies, a common practice is to save the current policy at the end of each iteration and perform an extensive evaluation to determine the optimal rule list. However, for complex tasks, this approach often requires numerous iterations, resulting in a large library of policies, which can be challenging to test comprehensively. To tackle this problem, we propose a policy competition mechanism. At its core, this mechanism involves conducting a small-scale evaluation after each iteration to eliminate underperforming policies, thus keeping the size of the policy library manageable.

From a theoretical perspective, the competition mechanism addresses the problem of optimization instability arising from approximate implementations. Specifically, the unbiased estimation of $U^{\pi'}-U^\pi$ is formally defined as:
\begin{equation}
    P(s,\pi(s),s')V_{t+1}^\pi(s')-P(s,a,s')V_{t+1}^\pi(s').
\end{equation}
When the sampling is extensive, both of these quantities tend to align in their distributions. Leveraging principles from Monte Carlo approximation, we approximate $P(s,\pi(s),s')V_{t+1}^\pi(s')-P(s,a,s')V_{t+1}^\pi(s')$ using $U^{\pi'}-U^\pi$. However, due to limited sampling, biases can creep into the approximation, leading to increased policy variance and unstable policy convergence during the optimization process. Therefore, we opt for approximating the outcomes of large-scale testing through continuous small-scale sampling.

In practical terms, we introduce a competition mechanism to manage the size of the policy library. Specifically, we maintain a policy library with a maximum size of $M$. After each iteration, the updated policy is saved at the end of the policy library, and the following actions are taken:
\begin{enumerate}
    \item Initially, a small-scale evaluation is conducted. For a policy $\pi_n$, where $n<N$, if $\pi_n$ outperforms policy $\pi_{n-1}$, we swap $\pi_n$ and $\pi_{n-1}$. In essence, policies that perform well in the test are ranked higher.
    \item When the total number of policies exceeds $N$, the least-performing policy is removed from the library.
\end{enumerate}
Throughout the iterative optimization process, policies with advantages gradually improve their rankings and remain in the competition. Ultimately, we consider the strategy with the highest rank to be the closest approximation to the optimal rule-based policy. By introducing the competitive mechanism during iterations, we reduce the perturbations during the iterative optimization process and lower the difficulty of testing.

\section{Evaluation}
In this section, we assess the performance of the FIPE approach on three representative tasks within the StarCraft II environment,i.e., $3m$, $2s\_vs\_1sc$, and $8m$. For each task, we employ a well-trained agent as the teacher.

\subsection{Experimental Setting}
\textbf{Baseline Selection.}
Recall that our goal is to take an episode from a target agent and extract interpretable rules. To achieve this, IPE methods fit episodes through a self-explainable model and then obtain interpretations directly from its interpretation rule list. We compare FIPE against two representative alternative methods: DAGGER~\cite{ross2011reduction} and VIPER~\cite{bastani2018verifiable}. These methods are chosen as baselines for comparison.

\textbf{Proposed Method.}
According to Section \ref{PolicyCompetition}, one of the key advancements of our proposed method is its ability to limit the number of models retained during the training process. For all tasks, we set the maximum number of models to 10 empirically. 

\textbf{Student Structure.}
We refer to the student model configurations used in existing methods and employ decision trees as the student model for all approaches. Additionally, we extend our validation to demonstrate the adaptability of our algorithm to other model architectures. In our case, we implement k-Nearest Neighbor and Support Vector Machine as alternative student model structures.

\subsection{Experimental Environment and Tasks}
The primary goal of the proposed method is to enhance the consistency of the interpretable student with the black-box teacher, thereby avoiding inconsistent explanation due to inconsistencies between the explanation and the original policy. To validate this claim, we conduct experiments in the challenging StarCraft II environment to evaluate the performance of our method and alternative approaches.

StarCraft II is renowned as an extremely challenging multi-agent cooperative game often used in reinforcement learning research. In this game, the red and blue forces confront each other. The agent controls the blue force, while the red force is controlled by a robust rules (with a difficulty level of 7). The objective of both the red and blue forces is to eliminate each other's units. At every step, each unit has access to a wealth of information, including (a) details about itself (e.g., health points, shield value, unit type); (b) information of nearby agents (e.g., distances to allies and enemies, relative coordinates, health points, unit types); and terrain characteristics (e.g., heights, maneuverability). Each unit can perform actions such as movement (in any of the 8 movable directions), attacking (based on the enemy's ID), stopping, or no option. Agents receive rewards for winning games or defeating enemy units.

In this study, we implement the proposed method and its alternatives on three challenging tasks: $3m$, $2s\_vs\_1sc$, and $8m$. Specifically, $3m$ is a small-scale task used to validate the feasibility of the methods on the StarCraft II platform. $2s\_vs\_1sc$ is a collaboration-intensive complex task where the agent controls two weak units against one strong enemy, making it essential to extract cooperative relationships. $8m$ is a complex task with a moderately large scale, where the agent controls eight marines, leading to a significant increase in input attribute dimensions. The primary challenge in the $8m$ map is dealing with high-dimensional inputs.

\subsection{Evaluation Metrics}\label{evaluationmetrics}
To evaluate the performance of the proposed method, we use the following metrics:

\textbf{Win rate.} We measure the average win rate of the tested models. Specifically, In StarCraft II tasks, the agent's goal is to win the game by defeating all opponents. We conduct 10 learning sessions for each method from scratch, and in each session, we calculate the average win rate over 20 episodes.

\textbf{Accumulated reward.} We measure the average accumulated reward of the tested models. Specifically, for each method under examination, we conduct 10 complete learning sessions, each comprising 20 testing rounds. The average accumulated rewards are computed by summing the rewards achieved throughout these testing rounds.

\textbf{Consistency.} The primary objective of the proposed method is to enhance the student's consistency with the teacher. To evaluate the consistency of the tested models, we validate the prediction accuracy of all tested models. Specifically, for each method, we perform 10 learning sessions from scratch, stopping the sampling when the data point quantity reaches 30,000. During the 20 interaction rounds, we calculate the average prediction accuracy compared to deep reinforcement learning policies. This average prediction accuracy is computed as $\mathbb{E}_{s\sim d^{\pi}}[\mathbb{I}(\pi(s),\tilde{\pi}(s))]$.

\textbf{Overall performance.} To provide a comprehensive evaluation of the policy extraction methods, we consider both the average reward from interactions with the environment and the consistency with the original policy. We design a composite metric by aligning the value ranges of reward and consistency. Formally, the composite metric is defined as:
\begin{equation}
    \Phi = \frac{1}{2N}\sum_{n=1}^N{[ norm(R_n)+A_n]},
\end{equation}
where $N$ is the number of test episodes, $R_n$ and $A_n$ represent the accumulated reward and consistency rate on episode $n$, respectively, and $norm(\cdot)$ is a normalization function.

\subsection{Performance and Stability}
\begin{figure}
    \centering\includegraphics[width=.9\linewidth]{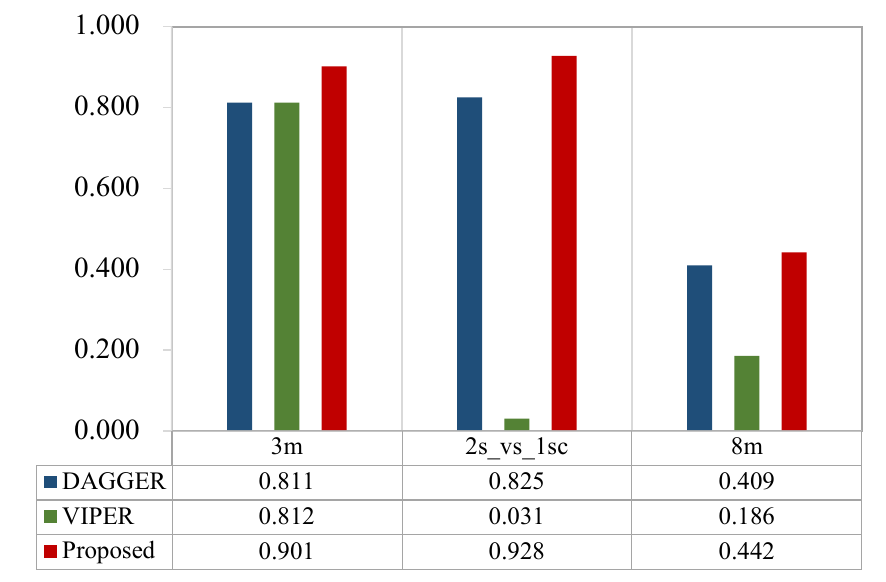}
    \caption{Comprehensive evaluation of the policy extraction methods. The comprehensive metric is consider both the average reward and the consistency of the methods, refer to the Section~\ref{evaluationmetrics} for formal representation.}
    \label{fig.OverallPerformance}
\end{figure}

In this section, we evaluate the performance and stability analysis of the proposed method. Specifically, we conduct a comprehensive optimization of the models and present the average results for each evaluation metric. Figure \ref{fig.Evaluation} (a) and Figure \ref{fig.OverallPerformance} showcases the experimental outcomes of FIPE in comparison to baseline alternative methods.

\subsubsection{Performance}
\begin{figure*}
    \centering
    \subfloat[]{\includegraphics[width=.5\linewidth]{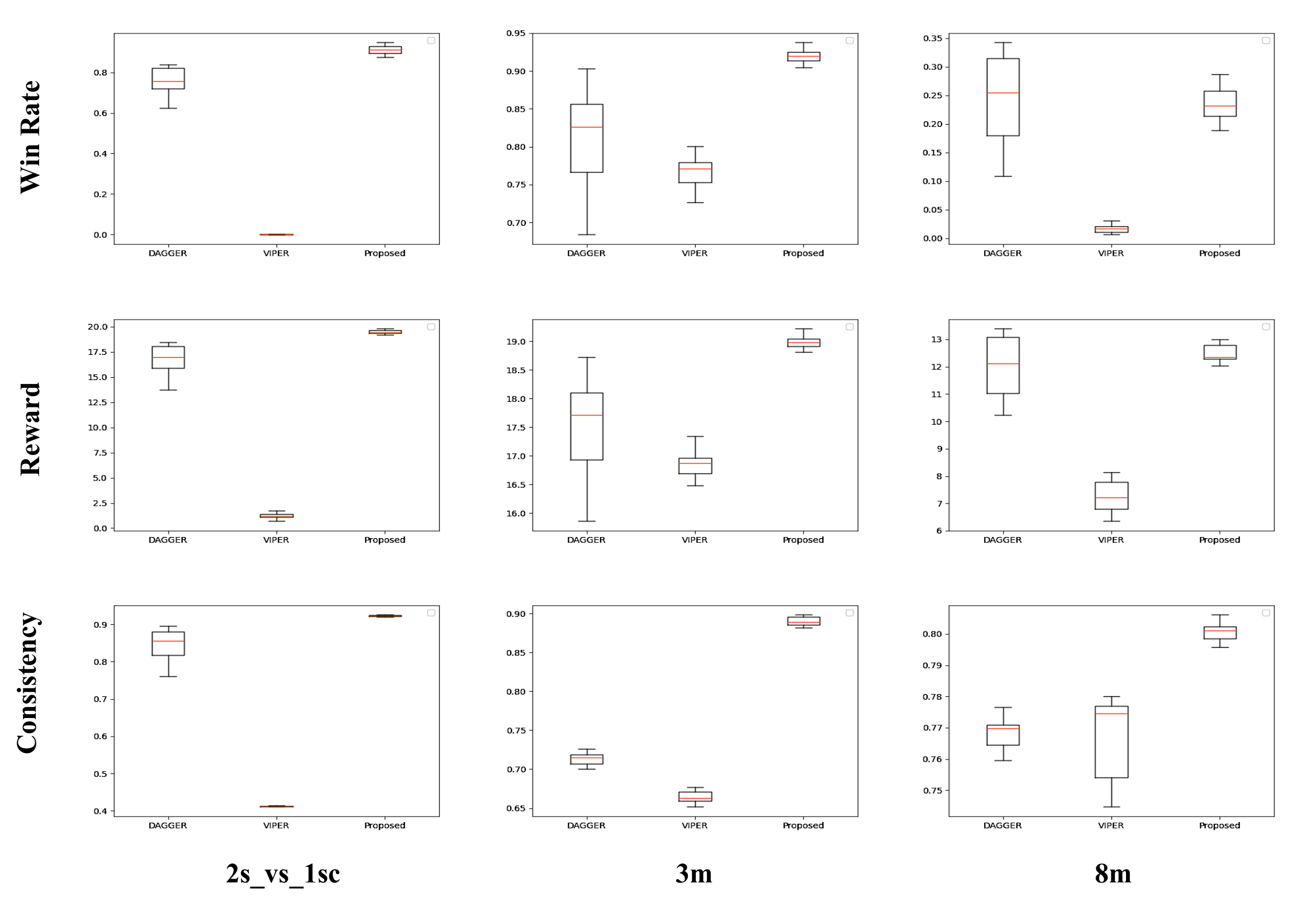}}
    \subfloat[]{\includegraphics[width=.5\linewidth]{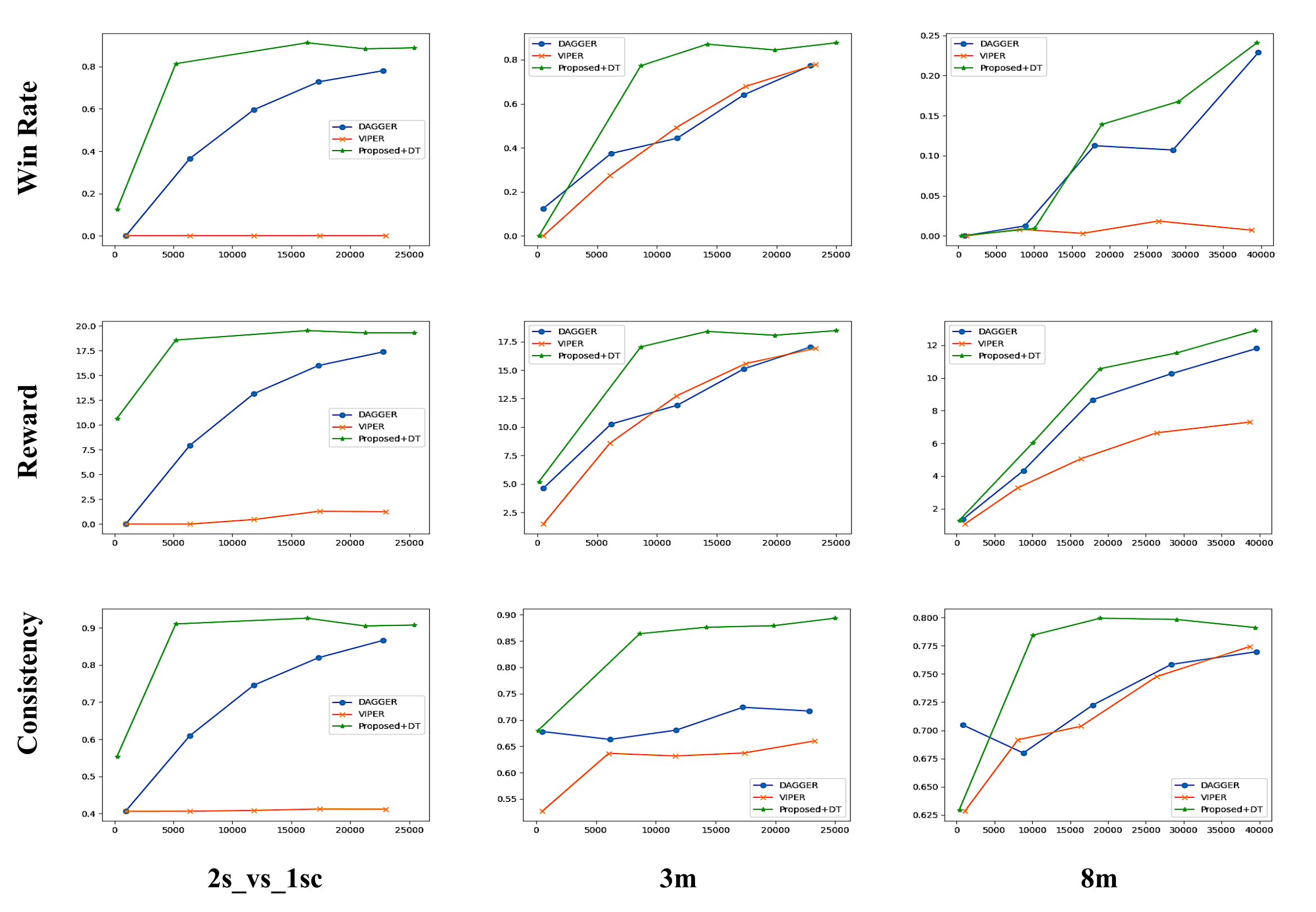}}
    \caption{
    (a) Evaluation of performance and stability. In each subgraph, the vertical axis represents the scale of the evaluation metric, and the horizontal axis is used to distinguish between different methods.(b) Evaluation of convergence rate. In each subgraph, the vertical axis represents the scale of the evaluation metric, while the horizontal axis represents the number of samples, signifying various training stages.
    }
    \label{fig.Evaluation}
\end{figure*}

As illustrated in Figure \ref{fig.Evaluation} (a), our proposed method consistently outperforms the baseline alternatives in terms of interactive performance and consistency across all three StarCraft II maps.

As for the $3m$ task, it's evident that all algorithms involved in the comparison demonstrate commendable performance. This suggests the applicability of policy extraction algorithms within the StarCraft II environment. As for the $2s\_vs\_1sc$ task, the proposed method significantly improves the consistency of policy fitting, effectively capturing the cooperative relationships between agents. As for the $8m$ task, extracting rules from such high-dimensional inputs is a challenging task for traditional decision trees. Despite none of the methods extracting complex behavior, the proposed method still outperforms the baselines.

Subsequently, as discussed in Section \ref{section_theory}, FIPE explicitly models fidelity, resulting in interpretable rule models that align with the original policy. As seen in Figure \ref{fig.Evaluation}-Consistency, the proposed method consistently outperforms the baselines in terms of consistency, highlighting its capability to enhance model consistency.

Finally, when considering both the win rate (Figure \ref{fig.Evaluation} (a) Win rate) and cumulative reward (Figure \ref{fig.Evaluation} (a) Reward), the proposed method achieves optimal performance in most scenarios. This aligns with the conclusions drawn in Section \ref{section_theory}, where we highlighted that the fidelity-inducing mechanism does not compromise the policy's average reward. Two key factors contributing to this are: (a) the fidelity-inducing term gradually approaches zero during the optimization process, and (b) inducing fidelity to the original policy can mitigate shortsightedness and enhance overall performance.

\subsubsection{Stability}
The stability of policy extraction methods directly impacts their robustness during deployment. As observed in the experiment results (depicted in Figure \ref{fig.Evaluation} (a)), the proposed method exhibits greater stability across various metrics, surpassing the baseline alternatives. We further conduct the following analysis:

First, existing methods, particularly VIPER, often struggle when applied to complex tasks. VIPER calculates sample advantages based on the action value function and tends to discard a significant portion of the samples. However, in complex adversarial scenarios within the StarCraft II platform, policies frequently need to handle intricate logic and abrupt situations. Since the advantage samples tend to be similar, VIPER's approach reduces the model's ability to generalize and discards a substantial number of valuable samples. Consequently, it produces unsatisfactory results for samples with low occurrence frequency, contributing to a decrease in overall stability.

Second, when compared to the baseline methods, the proposed approach consistently demonstrates a stability advantage across all metrics, with consistency being particularly notable. This reaffirms our previous discussion (Section \ref{section_theory}). Specifically, by emphasizing fidelity, the proposed method mitigates the shortsightedness of pursuing immediate rewards, leading to enhanced stability.

\subsubsection{Comprehensive Evaluation}
Figure \ref{fig.OverallPerformance} presents the results of a comprehensive metric that takes into account both the average reward and the consistency of the extracted model. Firstly, it's evident that the proposed method surpasses its competitors across all three tasks. This highlights the effectiveness of our approach in achieving a balance between reward maximization and model consistency. Secondly, the overall performance of the methods varies depending on the task's characteristics. Specifically, In the case of $3m$, which is relatively straightforward, all methods perform well. This indicates that simpler tasks are generally handled competently by these methods. $2s\_vs\_1sc$ is notably error-sensitive, where baseline methods tend to make irreparable mistakes. In contrast, the proposed method exhibits a higher level of robustness in this scenario. $8m$ poses significant challenges for all methods due to its complexity. However, the proposed method emerges as the best-performing approach even in this demanding task.

\subsection{Convergence Rate}
In order to evaluate the convergence rate of the proposed method, we present test results during the training process. Specifically, for $3m$ and $2s\_vs\_1sc$, we display the testing results from 1,000 to 25,000 samples. In the case of $8m$, we showcase testing results from 1,000 to 40,000 samples. These testing results are illustrated in Figure \ref{fig.Evaluation} (b).

First, the testing results (Figure \ref{fig.Evaluation} (b)) indicate that the convergence speed of policy extraction methods is linked to the number of agents in the environment. For instance, in $2s\_vs\_1sc$, where there are only three units on the map resulting in a relatively small state space, training progresses more swiftly. Conversely, in $3m$, where the total number of units on the map is six, convergence occurs at a slower pace. In $8m$, where the total number of units increases to 16, and the state space expands significantly, the training process becomes even more challenging.

Second, the proposed method showcases faster convergence in terms of win rate, average reward, and consistency compared to the baseline methods across all environments. In $3m$ and $2s\_vs\_1sc$, all metrics approach optimal performance with around 10,000 data points. Even in $8m$, where all methods face limitations, the proposed method achieves consistency convergence before reaching 10,000 samples. As discussed earlier, the primary aim of the proposed method is to enhance consistency with the original policy. The experiments convincingly demonstrate the method's efficiency in fitting policies with high consistency.


\begin{table}
    \centering
    \caption{Performance Ablation Experiment with Other Interpretable Target Models}
    \label{fig:c5_multi}
    \begin{tabular}{llcc}
        \hline
                                         & \textbf{Method} & \multicolumn{1}{l}{\textbf{Task: 3m}} & \multicolumn{1}{l}{\textbf{Task: 8m}} \\ \hline
        \textbf{Win Rate}   & DAGGER        & 0.772                                 & 0.228                                 \\
                                         & VIPER         & 0.778                                 & 0.007                                 \\
                                         & Proposed(DT)  & 0.877                                 & 0.240                                 \\
                                         & Proposed(KNN) & \textbf{0.944}                        & 0.643                                 \\
                                         & Proposed(SVM) & 0.862                                 & \textbf{0.788}                        \\ \hline
        \textbf{Reward}   & DAGGER        & 17.019                                & 11.799                                \\
                                         & VIPER         & 16.900                                & 7.295                                 \\
                                         & Proposed(DT)  & 18.465                                & 12.894                                \\
                                         & Proposed(KNN) & \textbf{19.316}                       & 17.133                                \\
                                         & Proposed(SVM) & 18.294                                & \textbf{18.354}                       \\ \hline
        \textbf{Consistency} & DAGGER        & 0.717                                 & 0.767                                 \\
                                         & VIPER         & 0.660                                 & 0.765                                 \\
                                         & Proposed(DT)  & 0.893                                 & \textbf{0.791}                                 \\
                                         & Proposed(KNN) & \textbf{0.923}                        & 0.774                                 \\
                                         & Proposed(SVM) & 0.903                                 & 0.769                        \\ \hline
    \end{tabular}
\end{table}

\subsection{Compatible with Different Self-Explainable Structures}
In this section, we assess the compatibility of the proposed algorithm with various interpretable model structures. In our previous experiments, we used decision trees as the student structure to maintain consistency with the baseline methods. However, we observed a decrease in the average performance of decision trees when dealing with complex tasks. In this section, we introduce k-nearest neighbors (KNN) and support vector machines (SVM) as alternative structures, demonstrating that the proposed method is adaptable to different interpretable model structures. The experiment results are presented in Table~\ref{fig:c5_multi}.

First, by replacing the decision tree with KNN or SVM structures, we observe significant improvements in both average rewards and consistency of the extracted policies. This phenomenon aligns with the discussions in this paper regarding the limited expressive power of decision trees. Moreover, it underscores the compatibility of the proposed framework with interpretable machine learning models within the existing supervised learning paradigms.

Second, in the case of $3m$ task, the win rate of the KNN policy extracted through deep reinforcement learning surpasses 94\%, and for $8m$ task, the extracted SVM policy achieves a win rate exceeding 78\%. It's worth noting that the teacher model in these experiments is Q-Mix. This demonstrates that the win rate of the extracted policy approaches that of the teacher model.

Existing IPE methods tend to focus on simpler tasks and avoid testing in complex platforms like StarCraft II, which involve adversarial and cooperative aspects, high-dimensional data, and intricate logic. In contrast, the proposed method enhances extraction performance by leveraging high-performance model structures, enabling its application in complex tasks.

\section{Discussion}

\textbf{Other games.}
Besides the two-party Markov game (i.e., StarCraft II) studied in this work, many other multiplayer Markov games also exhibit complexity, for example, real-time strategy games \cite{vinyals2019grandmaster,kurach2020google}, as well as extensive-form games ~\cite{silver2016mastering,silver2017mastering}. Regarding multiplayer Markov games, the associations between the episodes and final rewards will also be more sophisticated, necessitating a model with a high capacity to make accurate predictions. As part of future work, we will investigate methods to increase the capacity of our proposed model for these games, such as using self-interpretable models (e.g., soft decision tree ~\cite{frosst2017distilling}) that offer improved performance, even though this may reduce interpretability to some extent.

\textbf{Limitations and future works.}
Our work has limitations. First, while we achieve high consistency, the model's win rate is reduced, indicating that it cannot entirely replace deep reinforcement learning policies. As part of future work, we aim to enhance the trade-off between accuracy and interpretability. This could involve exploring advanced algorithms, maintaining interpretability while enhancing accuracy, and devising more sophisticated rule extraction strategies or novel objective functions to balance model performance and interpretability. Secondly, the effectiveness of our model is constrained by the choice of the interpretable student model, particularly in high-dimensional state spaces like the $8m$ task. Our future work will explore strategies to address challenges associated with high-dimensional state spaces, potentially including more efficient feature selection and dimensionality reduction methods, or novel policy extraction techniques tailored to such environments. Last, while we have evaluated the compatibility of our method with different self-explainable structures, there is room for improvement in leveraging the capabilities of these students. We will further refine the design of interpretable student models, considering how to blend self-interpretable models with DRL models for enhanced performance and a better balance between performance and interpretability. This may involve developing new model architectures or feature engineering techniques.

\section{Conclusion}
This paper introduces FIPE, a novel approach aimed at providing rule-based explanations for deep reinforcement learning policies. Technically, FIPE treats the DRL agent as a black-box, striving to express its decision-making behavior through a classic self-interpretable structure. In contrast to existing Interpretable Policy Extraction (IPE) methods, which often encounter explainability challenges due to inconsistencies between the extracted rule-based policy and the DRL model's actual decisions, FIPE addresses this issue by introducing a fidelity-inducing IPE method. Importantly, our method avoids the use of incomprehensible components while enabling IPE in complex tasks like StarCraft II.

\bibliographystyle{IEEEtran}
\bibliography{reference}

\begin{thebibliography}{10}
\providecommand{\url}[1]{#1}
\csname url@samestyle\endcsname
\providecommand{\newblock}{\relax}
\providecommand{\bibinfo}[2]{#2}
\providecommand{\BIBentrySTDinterwordspacing}{\spaceskip=0pt\relax}
\providecommand{\BIBentryALTinterwordstretchfactor}{4}
\providecommand{\BIBentryALTinterwordspacing}{\spaceskip=\fontdimen2\font plus
\BIBentryALTinterwordstretchfactor\fontdimen3\font minus
  \fontdimen4\font\relax}
\providecommand{\BIBforeignlanguage}[2]{{%
\expandafter\ifx\csname l@#1\endcsname\relax
\typeout{** WARNING: IEEEtran.bst: No hyphenation pattern has been}%
\typeout{** loaded for the language `#1'. Using the pattern for}%
\typeout{** the default language instead.}%
\else
\language=\csname l@#1\endcsname
\fi
#2}}
\providecommand{\BIBdecl}{\relax}
\BIBdecl

\bibitem{molnar2020interpretable}
C.~Molnar, \emph{Interpretable machine learning}.\hskip 1em plus 0.5em minus
  0.4em\relax Lulu. com, 2020.

\bibitem{vinyals2019grandmaster}
O.~Vinyals, I.~Babuschkin, W.~M. Czarnecki, M.~Mathieu, A.~Dudzik, J.~Chung,
  D.~H. Choi, R.~Powell, T.~Ewalds, P.~Georgiev \emph{et~al.}, ``Grandmaster
  level in starcraft ii using multi-agent reinforcement learning,''
  \emph{Nature}, vol. 575, no. 7782, pp. 350--354, 2019.

\bibitem{lanctot2017unified}
M.~Lanctot, V.~Zambaldi, A.~Gruslys, A.~Lazaridou, K.~Tuyls, J.~P{\'e}rolat,
  D.~Silver, and T.~Graepel, ``A unified game-theoretic approach to multiagent
  reinforcement learning,'' \emph{Advances in Neural Information Processing
  Systems (NeurIPS)}, vol.~30, 2017.

\bibitem{silver2017mastering}
D.~Silver, T.~Hubert, J.~Schrittwieser, I.~Antonoglou, M.~Lai, A.~Guez,
  M.~Lanctot, L.~Sifre, D.~Kumaran, T.~Graepel \emph{et~al.}, ``Mastering chess
  and shogi by self-play with a general reinforcement learning algorithm,''
  \emph{arXiv preprint arXiv:1712.01815}, 2017.

\bibitem{collins2005efficient}
S.~Collins, A.~Ruina, R.~Tedrake, and M.~Wisse, ``Efficient bipedal robots
  based on passive-dynamic walkers,'' \emph{Science}, vol. 307, no. 5712, pp.
  1082--1085, 2005.

\bibitem{johannink2019residual}
T.~Johannink, S.~Bahl, A.~Nair, J.~Luo, A.~Kumar, M.~Loskyll, J.~A. Ojea,
  E.~Solowjow, and S.~Levine, ``Residual reinforcement learning for robot
  control,'' in \emph{2019 International Conference on Robotics and Automation
  (ICRA)}.\hskip 1em plus 0.5em minus 0.4em\relax IEEE, 2019, pp. 6023--6029.

\bibitem{li2023deep}
C.~Li, P.~Zheng, Y.~Yin, B.~Wang, and L.~Wang, ``Deep reinforcement learning in
  smart manufacturing: A review and prospects,'' \emph{CIRP Journal of
  Manufacturing Science and Technology}, vol.~40, pp. 75--101, 2023.

\bibitem{bastani2016measuring}
O.~Bastani, Y.~Ioannou, L.~Lampropoulos, D.~Vytiniotis, A.~Nori, and
  A.~Criminisi, ``Measuring neural net robustness with constraints,''
  \emph{Advances in neural information processing systems}, vol.~29, 2016.

\bibitem{guo2021edge}
W.~Guo, X.~Wu, U.~Khan, and X.~Xing, ``Edge: Explaining deep reinforcement
  learning policies,'' \emph{Advances in Neural Information Processing
  Systems}, vol.~34, pp. 12\,222--12\,236, 2021.

\bibitem{romdhana2022deep}
A.~Romdhana, A.~Merlo, M.~Ceccato, and P.~Tonella, ``Deep reinforcement
  learning for black-box testing of android apps,'' \emph{ACM Transactions on
  Software Engineering and Methodology (TOSEM)}, vol.~31, no.~4, pp. 1--29,
  2022.

\bibitem{minh2022explainable}
D.~Minh, H.~X. Wang, Y.~F. Li, and T.~N. Nguyen, ``Explainable artificial
  intelligence: a comprehensive review,'' \emph{Artificial Intelligence
  Review}, pp. 1--66, 2022.

\bibitem{loh2022application}
H.~W. Loh, C.~P. Ooi, S.~Seoni, P.~D. Barua, F.~Molinari, and U.~R. Acharya,
  ``Application of explainable artificial intelligence for healthcare: A
  systematic review of the last decade (2011--2022),'' \emph{Computer Methods
  and Programs in Biomedicine}, p. 107161, 2022.

\bibitem{ross2011reduction}
S.~Ross, G.~Gordon, and D.~Bagnell, ``A reduction of imitation learning and
  structured prediction to no-regret online learning,'' in \emph{Proceedings of
  the fourteenth international conference on artificial intelligence and
  statistics}.\hskip 1em plus 0.5em minus 0.4em\relax JMLR Workshop and
  Conference Proceedings, 2011, pp. 627--635.

\bibitem{lee2019complementary}
J.~H. Lee, ``Complementary reinforcement learning towards explainable agents,''
  \emph{arXiv preprint arXiv:1901.00188}, 2019.

\bibitem{dahlin2020designing}
N.~Dahlin, K.~C. Kalagarla, N.~Naik, R.~Jain, and P.~Nuzzo, ``Designing
  interpretable approximations to deep reinforcement learning,'' \emph{arXiv
  preprint arXiv:2010.14785}, 2020.

\bibitem{bastani2018verifiable}
O.~Bastani, Y.~Pu, and A.~Solar-Lezama, ``Verifiable reinforcement learning via
  policy extraction,'' \emph{Advances in Neural Information Processing Systems
  (NeurIPS)}, vol.~31, 2018.

\bibitem{liu2019rethinking}
Z.~Liu, M.~Sun, T.~Zhou, G.~Huang, and T.~Darrell, ``Rethinking the value of
  network pruning,'' \emph{International Conference on Learning Representations
  (ICLR)}, 2019.

\bibitem{li2021neural}
Z.-H. Li, Y.~Yu, Y.~Chen, K.~Chen, Z.~Hu, and C.~Fan, ``Neural-to-tree policy
  distillation with policy improvement criterion,'' \emph{arXiv preprint
  arXiv:2108.06898}, 2021.

\bibitem{hussein2017imitation}
A.~Hussein, M.~M. Gaber, E.~Elyan, and C.~Jayne, ``Imitation learning: A survey
  of learning methods,'' \emph{ACM Computing Surveys (CSUR)}, vol.~50, no.~2,
  pp. 1--35, 2017.

\bibitem{ghasemipour2020divergence}
S.~K.~S. Ghasemipour, R.~Zemel, and S.~Gu, ``A divergence minimization
  perspective on imitation learning methods,'' in \emph{Conference on Robot
  Learning}.\hskip 1em plus 0.5em minus 0.4em\relax PMLR, 2020, pp. 1259--1277.

\bibitem{cha2020proxy}
H.~Cha, J.~Park, H.~Kim, M.~Bennis, and S.-L. Kim, ``Proxy experience replay:
  Federated distillation for distributed reinforcement learning,'' \emph{IEEE
  Intelligent Systems}, vol.~35, no.~4, pp. 94--101, 2020.

\bibitem{liu2019toward}
G.~Liu, O.~Schulte, W.~Zhu, and Q.~Li, ``Toward interpretable deep
  reinforcement learning with linear model u-trees,'' in \emph{Machine Learning
  and Knowledge Discovery in Databases: European Conference, ECML PKDD 2018,
  Dublin, Ireland, September 10--14, 2018, Proceedings, Part II 18}.\hskip 1em
  plus 0.5em minus 0.4em\relax Springer, 2019, pp. 414--429.

\bibitem{vasic2022moet}
M.~Vasi{\'c}, A.~Petrovi{\'c}, K.~Wang, M.~Nikoli{\'c}, R.~Singh, and
  S.~Khurshid, ``Moet: Mixture of expert trees and its application to
  verifiable reinforcement learning,'' \emph{Neural Networks}, vol. 151, pp.
  34--47, 2022.

\bibitem{coppens2019distilling}
Y.~Coppens, K.~Efthymiadis, T.~Lenaerts, A.~Now{\'e}, T.~Miller, R.~Weber, and
  D.~Magazzeni, ``Distilling deep reinforcement learning policies in soft
  decision trees,'' in \emph{Proceedings of the IJCAI 2019 workshop on
  explainable artificial intelligence}, 2019, pp. 1--6.

\bibitem{you2022interpretability}
Y.~You, J.~Sun, Y.~Guo, Y.~Tan, and J.~Jiang, ``Interpretability and accuracy
  trade-off in the modeling of belief rule-based systems,''
  \emph{Knowledge-Based Systems}, vol. 236, p. 107491, 2022.

\bibitem{tjoa2020survey}
E.~Tjoa and C.~Guan, ``A survey on explainable artificial intelligence (xai):
  Toward medical xai,'' \emph{IEEE transactions on neural networks and learning
  systems}, vol.~32, no.~11, pp. 4793--4813, 2020.

\bibitem{puiutta2020explainable}
E.~Puiutta and E.~M. Veith, ``Explainable reinforcement learning: A survey,''
  in \emph{International cross-domain conference for machine learning and
  knowledge extraction}.\hskip 1em plus 0.5em minus 0.4em\relax Springer, 2020,
  pp. 77--95.

\bibitem{rosenfeld2019explainability}
A.~Rosenfeld and A.~Richardson, ``Explainability in human--agent systems,''
  \emph{Autonomous Agents and Multi-Agent Systems}, vol.~33, pp. 673--705,
  2019.

\bibitem{silva2020optimization}
A.~Silva, M.~Gombolay, T.~Killian, I.~Jimenez, and S.-H. Son, ``Optimization
  methods for interpretable differentiable decision trees applied to
  reinforcement learning,'' in \emph{International conference on artificial
  intelligence and statistics}.\hskip 1em plus 0.5em minus 0.4em\relax PMLR,
  2020, pp. 1855--1865.

\bibitem{guo2023explainable}
Y.~Guo, J.~Campbell, S.~Stepputtis, R.~Li, D.~Hughes, F.~Fang, and K.~Sycara,
  ``Explainable action advising for multi-agent reinforcement learning,'' in
  \emph{2023 IEEE International Conference on Robotics and Automation
  (ICRA)}.\hskip 1em plus 0.5em minus 0.4em\relax IEEE, 2023, pp. 5515--5521.

\bibitem{wang2023explainable}
Y.-C. Wang, T.~Chen, and M.-C. Chiu, ``An explainable deep-learning approach
  for job cycle time prediction,'' \emph{Decision Analytics Journal}, vol.~6,
  p. 100153, 2023.

\bibitem{milani2022maviper}
S.~Milani, Z.~Zhang, N.~Topin, Z.~R. Shi, C.~Kamhoua, E.~E. Papalexakis, and
  F.~Fang, ``Maviper: Learning decision tree policies for interpretable
  multi-agent reinforcement learning,'' in \emph{Joint European Conference on
  Machine Learning and Knowledge Discovery in Databases}.\hskip 1em plus 0.5em
  minus 0.4em\relax Springer, 2022, pp. 251--266.

\bibitem{liu2023effective}
X.~Liu, S.~Liu, B.~An, Y.~Gao, S.~Yang, and W.~Li, ``Effective interpretable
  policy distillation via critical experience point identification,''
  \emph{IEEE Intelligent Systems}, 2023.

\bibitem{kurach2020google}
K.~Kurach, A.~Raichuk, P.~Sta{\'n}czyk, M.~Zaj{\k{a}}c, O.~Bachem, L.~Espeholt,
  C.~Riquelme, D.~Vincent, M.~Michalski, O.~Bousquet \emph{et~al.}, ``Google
  research football: A novel reinforcement learning environment,'' in
  \emph{Proceedings of the AAAI conference on artificial intelligence},
  vol.~34, no.~04, 2020, pp. 4501--4510.

\bibitem{silver2016mastering}
D.~Silver, A.~Huang, C.~J. Maddison, A.~Guez, L.~Sifre, G.~Van Den~Driessche,
  J.~Schrittwieser, I.~Antonoglou, V.~Panneershelvam, M.~Lanctot \emph{et~al.},
  ``Mastering the game of go with deep neural networks and tree search,''
  \emph{nature}, vol. 529, no. 7587, pp. 484--489, 2016.

\bibitem{frosst2017distilling}
N.~Frosst and G.~Hinton, ``Distilling a neural network into a soft decision
  tree,'' \emph{arXiv preprint arXiv:1711.09784}, 2017.

\end{thebibliography}

\newpage

\end{document}